\documentclass[accepted]{article}

\usepackage{ifthen}
\usepackage{microtype}
\usepackage{graphicx}
\usepackage{booktabs} 
\usepackage[normalem]{ulem} 
\usepackage{enumitem}
\usepackage{algorithm}
\usepackage[noend]{algorithmic}
\usepackage{amsmath,amssymb}
\usepackage{xspace}

\usepackage{hyperref}

\ifthenelse{\isundefined{\forreview}}{
  \usepackage[accepted]{sysml2019}
}{
  \usepackage{sysml2019}
}

\usepackage{caption}
\newcommand{\Section}[1]{\vspace{-2pt}\section{#1}\vspace{-1pt}}
\newcommand{\SubSection}[1]{\vspace{-2pt}\subsection{#1}\vspace{-1pt}}
\newcommand{\Paragraph}[1]{\vspace{-2pt}\paragraph{#1}\vspace{-1pt}}

\newcommand{\SUB}[1]{\ENSURE \hspace{-0.15in} \textbf{#1}}

\newcommand{\grad}{\triangledown}
\newcommand{\algfont}[1]{\texttt{#1}}
\newcommand{\fedavglong}{\algfont{FederatedAveraging}\xspace}

\newcommand{\loss}{\ell}
\renewcommand{\algorithmicensure}{}

\definecolor{darkgreen}{rgb}{0,0.4,0.0}

\sysmltitlerunning{Towards Federated Learning at Scale: System Design}

\begin{document}

\twocolumn[
\sysmltitle{Towards Federated Learning at Scale: System Design}



\sysmlsetsymbol{equal}{*}

\begin{sysmlauthorlist}
\sysmlauthor{Keith Bonawitz}{google}
\sysmlauthor{Hubert Eichner}{google}
\sysmlauthor{Wolfgang Grieskamp}{google}
\sysmlauthor{Dzmitry Huba}{google}
\sysmlauthor{Alex Ingerman}{google}
\sysmlauthor{Vladimir Ivanov}{google}
\sysmlauthor{Chlo\'e Kiddon}{google}
\sysmlauthor{Jakub Kone{\v{c}}n{\'y}}{google}
\sysmlauthor{Stefano Mazzocchi}{google}
\sysmlauthor{H. Brendan McMahan}{google}
\sysmlauthor{Timon Van Overveldt}{google}
\sysmlauthor{David Petrou}{google}
\sysmlauthor{Daniel Ramage}{google}
\sysmlauthor{Jason Roselander}{google}
\end{sysmlauthorlist}

\sysmlaffiliation{google}{Google Inc., Mountain View, CA, USA}

\sysmlcorrespondingauthor{Wolfgang Grieskamp}{wgg@google.com}
\sysmlcorrespondingauthor{Vladimir Ivanov}{vlivan@google.com}
\sysmlcorrespondingauthor{Brendan McMahan}{mcmahan@google.com}

\sysmlkeywords{Machine Learning, SysML, Federated Learning}

\vskip 10pt

\begin{abstract}
Federated Learning is a distributed machine learning approach which
enables model training on a large corpus of decentralized data.  We have
built a scalable production system for Federated Learning in the
domain of mobile devices, based on TensorFlow. In this paper, we describe
the resulting high-level design, sketch some of the challenges
and their solutions, and touch upon the open problems and
future directions.
\end{abstract}
]


\printAffiliationsAndNotice{}

\ifthenelse{\isundefined{\forreview}}{
  \newcommand{\ackn}{
    Galen Andrew,
    Blaise Ag{\"u}era y Arcas,
    Sean Augenstein,
    Dave Bacon,
    Fran\c{c}oise Beaufays,
    Amlan Chakraborty,
    Arlie Davis,
    Stefan Dierauf,
    Randy Dodgen,
    Emily Glanz,
    Shiyu Hu,
    Ben Kreuter,
    Eric Loewenthal,
    Antonio Marcedone,
    Jason Hunter,
    Krzysztof Ostrowski,
    Sarvar Patel,
    Peter Kairouz,
    Kanishka Rao,
    Michael Reneer,
    Aaron Segal,
    Karn Seth,
    Wei Huang,
    Nicholas Kong,
    Haicheng Sun,
    Vivian Lee,
    Tim Yang,
    Yuanbo Zhang%
  }
}{
  \newcommand{\ackn}{Anonymous Contributors}
}

\Section{Introduction}
\label{intro}

Federated Learning (FL) \citep{FL4} is a distributed machine learning
approach which enables training on a large corpus of decentralized
data residing on devices like mobile phones.  FL is one instance of
the more general approach of ``bringing the code to the data, instead
of the data to the code'' and addresses the fundamental problems of
privacy, ownership, and locality of data. The general description of
FL has been given by \citet{FL_BLOG}, and its theory has been explored
in \citet{FL3, FL4, mcmahan18dplm}.

A basic design decision for a Federated Learning infrastructure is
whether to focus on asynchronous or synchronous training
algorithms. While much successful work on deep learning has used
asynchronous training, e.g., \citet{dean12large}, recently there has
been a consistent trend towards synchronous large batch training, even
in the data center \citep{goyal17onehour,smith18dont}. The Federated
Averaging algorithm of \citet{FL4} takes a similar approach. Further,
several approaches to enhancing privacy guarantees for FL, including
differential privacy \citep{mcmahan18dplm} and Secure Aggregation
\citep{SECAGG}, essentially require some notion of synchronization on
a fixed set of devices, so that the server side of the learning
algorithm only consumes a simple aggregate of the updates from many
users. For all these reasons, we chose to focus on support for
synchronous rounds, while mitigating potential synchronization
overhead via several techniques we describe subsequently. Our system
is thus amenable to running large-batch SGD-style algorithms as well
as Federated Averaging, the primary algorithm we run in
production; pseudo-code is given in Appendix \ref{app:fedavg} for completeness.

In this paper, we report on a system design for such algorithms in the
domain of mobile phones (Android).  This work is still in an early
stage, and we do not have all problems solved, nor are we able to give
a comprehensive discussion of all required components. Rather, we
attempt to sketch the major components of the system, describe the
challenges, and identify the open issues, in the hope that this will
be useful to spark further systems research.

Our system enables one to train a deep neural network, using
TensorFlow \citep{TF}, on data stored on the phone which will never
leave the device. The weights are combined in the cloud with Federated
Averaging, constructing a global model which is pushed back to phones
for inference. An implementation of Secure Aggregation \citep{SECAGG}
ensures that on a global level individual updates from phones are
uninspectable.  The system has been applied in large scale
applications, for instance in the realm of a phone keyboard.

Our work addresses numerous practical issues: device availability
that correlates with the local data distribution in complex ways
(e.g., time zone dependency); unreliable device connectivity and
interrupted execution; orchestration of lock-step execution across
devices with varying availability; and limited device storage and
compute resources. These issues are addressed at the communication
protocol, device, and server levels. We have reached a state of
maturity sufficient to deploy the system in production and solve
applied learning problems over tens of millions of real-world devices;
we anticipate uses where the number of devices reaches billions.

\begin{figure*}[h!]
  \centering
  \includegraphics[width=\textwidth]{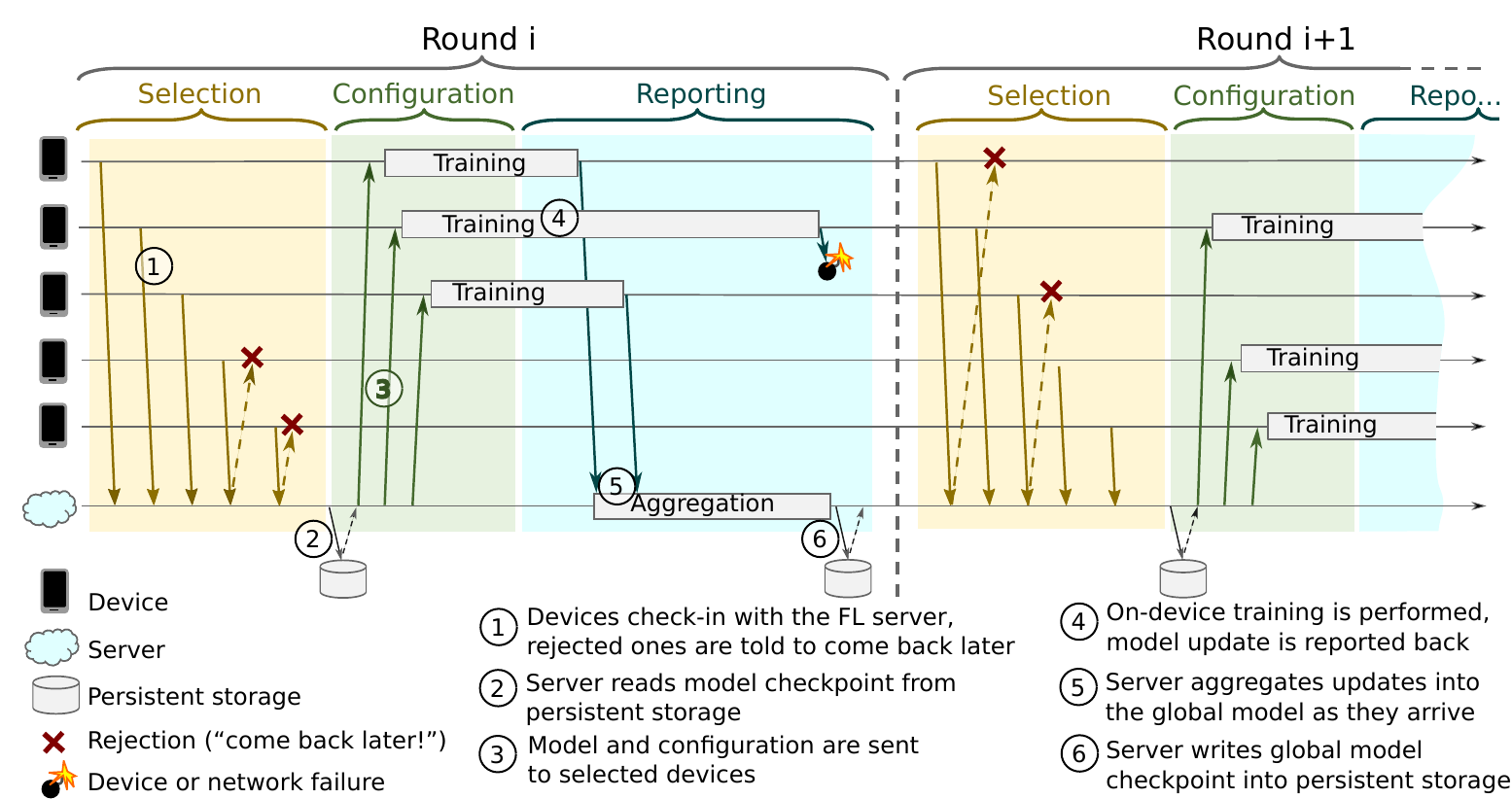}
  \caption{Federated Learning Protocol}
  \label{fig:FLProtocol}
\end{figure*}

\Section{Protocol}
\label{sec:Protocol}

To understand the system architecture, it is best to start from the
network protocol.

\SubSection{Basic Notions}
\label{sec:Notions}

The participants in the protocol are \textit{devices} (currently Android
phones) and the \textit{FL server}, which is a cloud-based distributed
service.
Devices announce to the server that they are ready to run an \textit{FL
 task} for a given \textit{FL population}. An FL population is specified
by a globally unique name which identifies the learning problem, or
application, which is worked upon. An FL task is a specific computation
for an FL population, such as training to be performed with given
hyperparameters, or evaluation of trained models on local device data.

From the potential tens of thousands of devices announcing
availability to the server during a certain time window, the server
selects a subset of typically a few hundred which are invited to work
on a specific FL task (we discuss the reason for this subsetting in
Sec.~\ref{sec:Phases}). We call this rendezvous between devices and
server a \textit{round}. Devices stay connected to the server for the
duration of the round.

The server tells the selected devices what computation to run with an
\textit{FL plan}, a data structure that includes a TensorFlow graph
and instructions for how to execute it.  Once a round is established,
the server next sends to each participant the current global model
parameters and any other necessary state as an \textit{FL checkpoint}
(essentially the serialized state of a TensorFlow session).  Each
participant then performs a local computation based on the global
state and its local dataset, and sends an update in the form of an FL
checkpoint back to the server. The server incorporates these updates
into its global state, and the process repeats.

\SubSection{Phases}
\label{sec:Phases}

The communication protocol enables devices to advance the global,
singleton model of an FL population between rounds where each round
consists of the three phases shown in Fig.~\ref{fig:FLProtocol}.
For simplicity, the description below does not include Secure
Aggregation, which is described in Sec.~\ref{sec:SecAgg}. Note that
even in the absence of Secure Aggregation, all network traffic is
encrypted on the wire.

\begin{description}[style=unboxed,leftmargin=0pt]
\item[Selection] Periodically, devices that meet the eligibility
  criteria (e.g., charging and connected to an unmetered network; see
  Sec.~\ref{sec:device}) check in to the server by opening a
  bidirectional stream.  The stream is used to track liveness and
  orchestrate multi-step communication. The server selects a subset of
  connected devices based on certain goals like the optimal number of
  participating devices (typically a few hundred devices participate
  in each round). If a device
  is not selected for participation, the server responds with
  instructions to reconnect at a later point in time.\footnote{In the
    current implementation, selection is done by simple reservoir
    sampling, but the protocol is amenable to more sophisticated
    methods which address selection bias.}
\item[Configuration] The server is configured based on the
  aggregation mechanism selected (e.g., simple or Secure Aggregation)
  for the selected devices. The server sends the FL plan and an FL checkpoint
  with the global model to each of the devices.
\item[Reporting] The server waits for the participating devices to
  report updates. As updates are received, the server aggregates them
  using Federated Averaging and instructs the reporting devices when
  to reconnect (see also Sec.~\ref{sec:PaceSteering}). If enough
  devices report in time, the round will be successfully completed
  and the server will update its global model, otherwise the round is
  abandoned.
\end{description}

As seen in Fig.~\ref{fig:FLProtocol}, straggling devices which do
not report back in time or do not react on configuration by the
server will simply be ignored. The protocol has a certain tolerance
for such drop-outs which is configurable per FL task.

The selection and reporting phases are specified by a set of
parameters which spawn flexible time windows. For example, for the
selection phase the server considers a device participant goal count,
a timeout, and a minimal percentage of the goal count which is
required to run the round. The selection phase lasts until the goal
count is reached or a timeout occurs; in the latter case, the round
will be started or abandoned depending on whether the minimal goal
count has been reached.

\SubSection{Pace Steering}
\label{sec:PaceSteering}

Pace steering is a flow control mechanism regulating the
pattern of device connections. It enables the FL server both to scale
down to handle small FL populations as well to scale up to very large
FL populations.

Pace steering is based on the simple mechanism of the server suggesting
to the device the optimum time window to reconnect. The device attempts to
respect this, modulo its eligibility.

In the case of small FL populations, pace steering is used to ensure
that a sufficient number of devices connect to the server
simultaneously. This is important both for the rate of task progress
and for the security properties of the Secure Aggregation
protocol. The server uses a stateless probabilistic algorithm
requiring no additional device/server communication to suggest reconnection
times to rejected devices so that subsequent checkins are likely to
arrive contemporaneously.

For large FL populations, pace steering is used to randomize device
check-in times, avoiding the ``thundering herd'' problem, and
instructing devices to connect as frequently as needed to run all
scheduled FL tasks, but not more.

Pace steering also takes into account the diurnal oscillation in the
number of active devices, and is able to adjust the time window
accordingly, avoiding excessive activity during peak hours and without
hurting FL performance during other times of the day.

\Section{Device}
\label{sec:device}

\begin{figure}[htb]
  \includegraphics[width=\columnwidth]{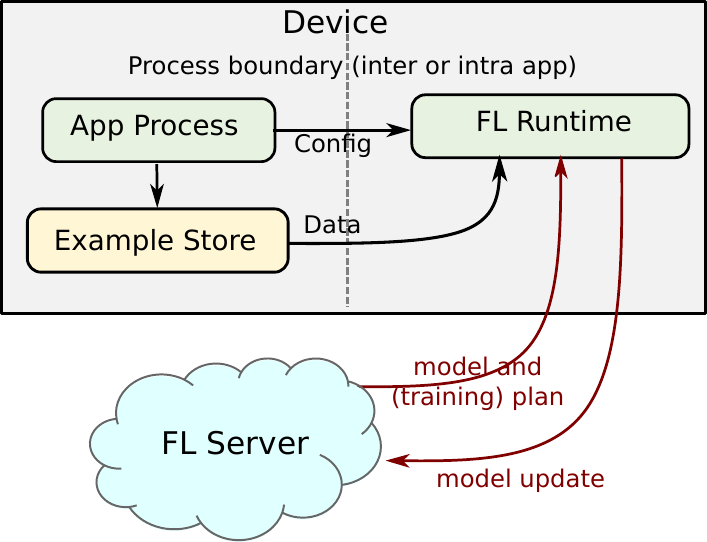}
  \caption{Device Architecture}
  \label{fig:DeviceArch}
\end{figure}

This section describes the software architecture running on a device
participating in FL. This describes our Android implementation but note
that the architectural choices made here are not particularly
platform-specific.

The device's first responsibility in on-device learning is to maintain
a repository of locally collected data for model training and
evaluation. Applications are responsible for making their data
available to the FL runtime as an \textit{example store} by
implementing an API we provide. An application's example store might,
for example, be an SQLite database recording action suggestions shown
to the user and whether or not those suggestions were accepted. We
recommend that applications limit the total storage footprint of their
example stores, and automatically remove old data after a
pre-designated expiration time, where appropriate. We provide
utilities to make these tasks easy. Data stored on devices may be
vulnerable to threats like malware or physical disassembly of the
phone, so we recommend that applications follow the best practices for
on-device data security, including ensuring that data is encrypted at
rest in the platform-recommended manner.

The FL runtime, when provided a task by the FL server, accesses an appropriate
example store to compute model updates, or evaluate model quality on held out
data. Fig.~\ref{fig:DeviceArch} shows the relationship between the
example store and the FL runtime. Control flow consists of the following steps:

\begin{description}[style=unboxed,leftmargin=0cm]
\item[Programmatic Configuration] An application configures the FL
  runtime by providing an FL population name and registering its
  example stores.  This schedules a periodic FL runtime job using
  Android's JobScheduler.  Possibly the most important requirement for
  training machine learning (ML) models on end users' devices is to
  avoid any negative impact on the user experience, data usage, or
  battery life. The FL runtime requests that the job scheduler only
  invoke the job when the phone is idle, charging, and connected to an
  unmetered network such as WiFi. Once started, the FL runtime will abort,
  freeing the allocated resources, if these conditions are no longer met.

\item[Job Invocation] Upon invocation by the job scheduler in a
  separate process, the FL runtime contacts the FL server to announce
  that it is ready to run tasks for the given FL population. The
  server decides whether any FL tasks are available for the population
  and will either return an FL plan or a suggested time to check in
  later.

\item[Task Execution] If the device has been selected, the FL runtime receives
  the FL plan, queries the app's example store for data requested by the plan,
  and computes plan-determined model updates and metrics.

\item[Reporting] After FL plan execution, the FL runtime reports computed
  updates and metrics to the server and cleans up any temporary resources.
\end{description}

As already mentioned, FL plans are not specialized to training, but can also
encode \textit{evaluation} tasks - computing quality metrics from
held out data that wasn't used for training, analogous to the
validation step in data center training.

Our design enables the FL runtime to either run within the application that
configured it or in a centralized service hosted in another app. Choosing
between these two requires minimal code changes. Communication between the
application, the FL runtime, and the application's example store as depicted in
Fig.~\ref{fig:DeviceArch} is implemented via Android's AIDL IPC mechanism, which
works both within a single app and across apps.

\Paragraph{Multi-Tenancy}

Our implementation provides a \textit{multi-tenant} architecture,
supporting training of multiple FL populations in the same app (or service).
This allows for coordination between multiple training activities, avoiding
the device being overloaded by many simultaneous training sessions at once.

\Paragraph{Attestation}

We want devices to participate in FL anonymously, which excludes the
possibility of authenticating them via a user identity.
Without verifying user identity, we need to protect against attacks to
influence the FL result from non-genuine devices. We do so by using Android's
remote attestation mechanism \citep{ATTESTATION}, which helps to ensure
that only genuine devices and applications participate in FL, and
gives us some protection against data poisoning \citep{BACKDOOR_FL}
via compromised devices. Other forms of model manipulation -- such as content
farms using uncompromised phones to steer a model -- are also potential
areas of concern that we do not address in the scope of this paper.

\Section{Server}
\label{server}

The design of the FL server is driven by the necessity to operate
over many orders of magnitude of population sizes and other
dimensions. The server must work with FL populations whose sizes range
from tens of devices (during development) to hundreds of millions, and
be able to process rounds with participant count ranging from tens of
devices to tens of thousands. Also, the size of the updates collected
and communicated during each round can range in size from kilobytes to
tens of megabytes. Finally, the amount of traffic coming into or out
of any given geographic region can vary dramatically over a day based
on when devices are idle and charging. This section details the design
of the FL server infrastructure given these requirements.

\SubSection{Actor Model}

The FL server is designed around the \textit{Actor Programming Model}
\citep{ACTORS}. Actors are universal primitives of concurrent computation which
use message passing as the sole communication mechanism.

Each actor handles a stream of messages/events strictly sequentially,
leading to a simple programming model. Running multiple instances of
actors of the same type allows a natural scaling to large number of
processors/machines.  In response to a message, an actor can make
local decisions, send messages to other actors, or create more actors
dynamically. Depending on the function and scalability requirements, actor
instances can be co-located on the same process/machine or distributed
across data centers in multiple geographic regions, using either
explicit or automatic configuration mechanisms. Creating and placing
fine-grained ephemeral instances of actors just for the duration of a
given FL task enables dynamic resource management and load-balancing decisions.

\SubSection{Architecture}

\begin{figure}[htb]
  \centering
  \includegraphics[width=0.9\columnwidth]{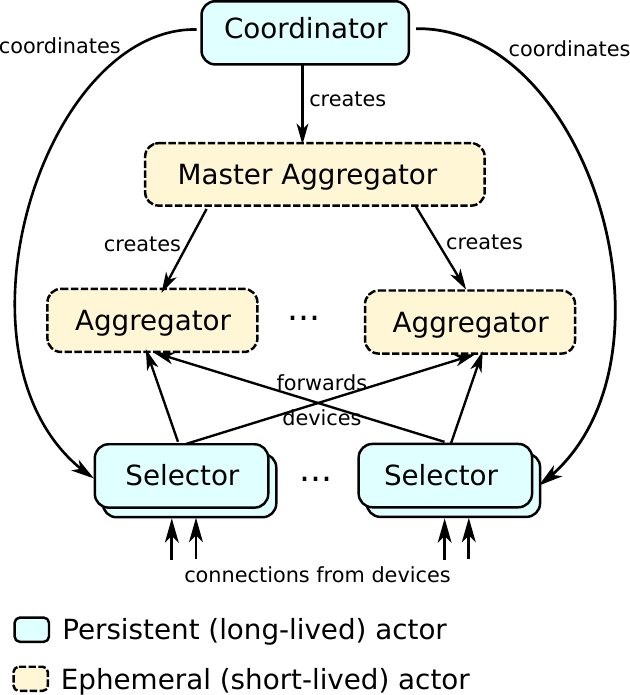}
  \caption{Actors in the FL Server Architecture}
  \label{fig:server}
\end{figure}

The main actors in the system are shown in Fig.~\ref{fig:server}.

\vspace{-10pt}
\begin{description}[style=unboxed,leftmargin=0pt]
\item[Coordinators] are the top-level actors which enable global
  synchronization and advancing rounds in lockstep. There are multiple
  Coordinators, and each one is responsible for an FL population of
  devices. A Coordinator registers its address and the FL population it
  manages in a shared locking service, so there is always a single
  owner for every FL population which is reachable by other actors in the
  system, notably the Selectors. The Coordinator receives information
  about how many devices are connected to each Selector and instructs them
  how many devices to accept for participation, based on which FL tasks
  are scheduled. Coordinators spawn Master Aggregators to manage the
  rounds of each FL task.
\item[Selectors] are responsible for accepting and forwarding device
  connections. They periodically receive information from the
  Coordinator about how many devices are needed for each FL
  population, which they use to make local decisions about whether or
  not to accept each device.  After the Master Aggregator and set of
  Aggregators are spawned, the Coordinator instructs the Selectors to
  forward a subset of its connected devices to the Aggregators,
  allowing the Coordinator to efficiently allocate devices to FL tasks
  regardless of how many devices are available. The approach also
  allows the Selectors to be globally distributed (close to devices)
  and limit communication with the remote Coordinator.
\item[Master Aggregators] manage the rounds of each FL task. In order
  to scale with the number of devices and update size, they make
  dynamic decisions to spawn one or more \textbf{Aggregators} to which
  work is delegated.
\end{description}

\vspace{-1ex}

No information for a round is written to persistent storage until it is fully
aggregated by the Master Aggregator. Specifically, all actors keep their state
in memory and are ephemeral. Ephemeral actors improve scalability by removing
the latency normally incurred by distributed storage. In-memory aggregation also
removes the possibility of attacks within the data center that target persistent
logs of per-device updates, because no such logs exist.

\SubSection{Pipelining}

While Selection, Configuration and Reporting phases of a round
(Sec.~\ref{sec:Protocol}) are sequential, the Selection phase doesn't
depend on any input from a previous round. This enables latency
optimization by running the Selection phase of the next round of the
protocol in parallel with the Configuration/Reporting phases of a previous
round. Our system architecture enables such pipelining without adding
extra complexity, as parallelism is achieved simply by the virtue of
Selector actors running the selection process continuously.

\SubSection{Failure Modes}

In all failure cases the system will continue to make progress, either
by completing the current round or restarting from the results of the
previously committed round. In many cases, the loss of an actor will
not prevent the round from succeeding. For example, if an Aggregator
or Selector crashes, only the devices connected to that actor will be
lost. If the Master Aggregator fails, the current round of the FL task
it manages will fail, but will then be restarted by the
Coordinator. Finally, if the Coordinator dies, the Selector
layer will detect this and respawn it. Because the Coordinators are
registered in a shared locking service, this will happen exactly once.

\Section{Analytics}
There are many factors and failsafes in the interaction between
devices and servers. Moreover, much of the platform activity happens on
devices that we neither control nor have access to.

For this reason, we rely on analytics to understand what is actually
going on in the field, and monitor devices' health statistics.  On the
device side we perform computation-intensive operations, and must
avoid wasting the phone's battery or bandwidth, or degrading the
performance of the phone. To ensure this, we log several activity and
health parameters to the cloud. For example: the device state in which
training was activated, how often and how long it ran, how much memory
it used, which errors where detected, which phone model / OS / FL
runtime version was used, and so on.  These log entries do not contain
any personally identifiable information (PII). They are aggregated and
presented in dashboards to be analyzed, and fed into automatic
time-series monitors that trigger alerts on substantial deviations.

We also log an event for every state in a training round, and use
these logs to generate ASCII visualizations of the sequence of state
transitions happening across all devices (see
Table~\ref{tab:TrainingRoundVisualizations} in the appendix). We chart
counts of these sequence visualizations in our dashboards, which
allows us to quickly distinguish between different types of
issues. For example, the sequence ``checking in, downloaded plan,
started training, ended training, starting upload, error'' is
visualized as ``\algfont{-v[]+*}\xspace'', while the shorter sequence
``checking in, downloaded plan, started training, error'' is
``\algfont{-v[*}\xspace''. The first indicates that a model trained
  successfully but the results upload failed (a network issue),
  whereas the second indicates that a training round failed right
  after loading the model (a model issue).

Server side, we similarly collect information such as how many
devices where accepted and rejected per training round, the timing of
the various phases of the round, throughput in terms of uploaded and
downloaded data, errors, and so on.

Since the platform's deployment, we have relied on the analytics layer
repeatedly to discover issues and verify that they were resolved. Some
of the incidents we discovered were device health related, for example
discovering that training was happening when it shouldn't have, while
others were functional, for example discovering that the drop out
rates of training participants were much higher than expected.

Federated training does not impact the user experience,
so both device and server \textit{functional} failures do not have an
immediate negative impact. But failures to operate properly could have
secondary consequences leading to utility degradation of the device. Device
utility to the user is mission critical, and degradations are difficult
to pinpoint and easy to wrongly diagnose.
Using accurate analytics to prevent federated training
from negatively impacting the device's utility
to the user accounts for a substantial part of our engineering
and risk mitigation costs.

\Section{Secure Aggregation}
\label{sec:SecAgg}

\citet{SECAGG} introduced \textit{Secure Aggregation}, a Secure
Multi-Party Computation protocol that uses encryption to make
individual devices' updates uninspectable by a server, instead only
revealing the sum after a sufficient number of updates have been
received. We can deploy Secure Aggregation as a privacy enhancement to
the FL service that protects against additional threats within the
data center by ensuring that individual devices' updates remain
encrypted even in-memory. Formally, Secure Aggregation protects from
``honest but curious'' attackers that may have access to the memory of
Aggregator instances. Importantly, the only aggregates needed for
model evaluation, SGD, or Federated Averaging are sums (e.g.,
$\bar{w}_t$ and $\bar{n}_t$ in Appendix~\ref{alg:fedavg}).\footnote{
  It is important to note that the
  goal of our system is to provide the tools to build privacy
  preserving applications. Privacy is enhanced by the ephemeral and
  focused nature of the FL updates, and can be further augmented with
  Secure Aggregation and/or differential privacy --- e.g., the
  techniques of \citet{mcmahan18dplm} are currently implemented.
  However, while the platform is designed to support a variety of
  privacy-enhancing technologies, stating specific privacy guarantees
  depends on the details of the application and the details of how
  these technologies are used; such a discussion is beyond the scope
  of the current work.}

Secure Aggregation is a four-round interactive protocol optionally
enabled during the reporting phase of a given FL round.  In each
protocol round, the server gathers messages from all devices in the FL
round, then uses the set of device messages to compute an independent
response to return to each device.  The protocol is designed to be
robust to a significant fraction of devices dropping out before the
protocol is complete. The first two rounds constitute a Prepare phase,
in which shared secrets are established and during which devices who
drop out will not have their updates included in the final
aggregation.  The third round constitutes a Commit phase, during which
devices upload cryptographically masked model updates and the server
accumulates a sum of the masked updates.  All devices who complete
this round will have their model update included in the protocol's
final aggregate update, or else the entire aggregation will fail.  The
last round of the protocol constitutes a Finalization phase, during
which devices reveal sufficient cryptographic secrets to allow the
server to unmask the aggregated model update.  Not all committed
devices are required to complete this round; so long as a sufficient
number of the devices who started to protocol survive through the
Finalization phase, the entire protocol succeeds.

Several costs for Secure Aggregation grow quadratically with the
number of users, most notably the computational cost for the
server. In practice, this limits the maximum size of a Secure
Aggregation to hundreds of users. So as
not to constrain the number of users that may participate in each
round of federated computation, we run an instance of Secure
Aggregation on each Aggregator actor (see Fig.~\ref{fig:server}) to
aggregate inputs from that Aggregator's devices into an intermediate
sum; FL tasks define a parameter $k$ so that all updates
are securely aggregated over groups of size at least $k$. The Master
Aggregator then further aggregates the intermediate aggregators'
results into a final aggregate for the round, without Secure
Aggregation.

\Section{Tools and Workflow}

\begin{figure*}[htb]
  \centering
  \includegraphics[width=\textwidth]{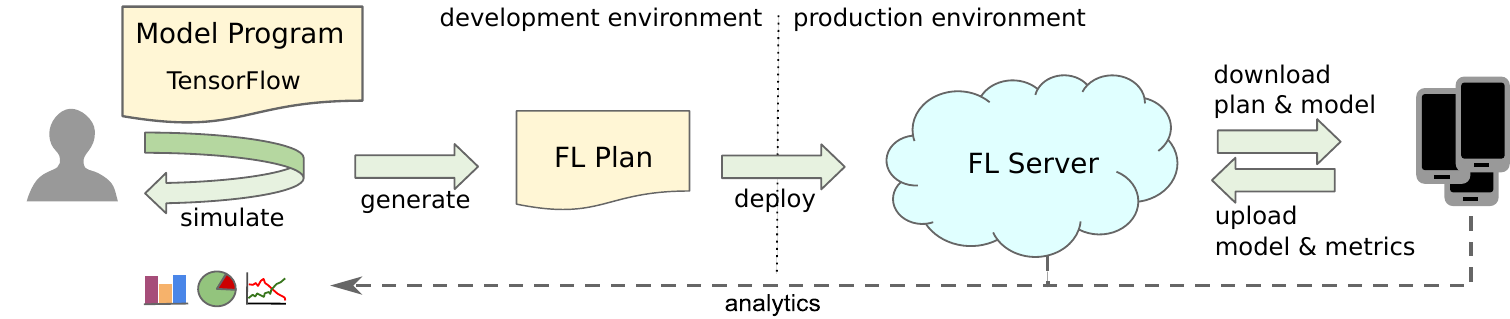}
  \caption{Model Engineer Workflow}
  \label{fig:Workflow}
  \vspace{-10pt}
\end{figure*}

Compared to the standard model engineer workflows on centrally collected
data, on-device training poses multiple novel challenges. First, individual
training examples are not directly inspectable, requiring tooling to work with
proxy data in testing and simulation (Sec.~\ref{sec:Modeling}). Second, models
cannot be run interactively but must instead be compiled into FL plans to be
deployed via the FL server (Sec.~\ref{sec:PlanGen}). Finally, because FL plans
run on real devices, model resource consumption and runtime compatibility must
be verified automatically by the infrastructure (Sec.~\ref{sec:Versioning}).

The primary developer surface of model engineers working with the FL system
is a set of Python interfaces and tools to define, test, and deploy
TensorFlow-based FL tasks to the fleet of mobile devices via the FL server.
The workflow of a model engineer for FL is depicted in Fig.~\ref{fig:Workflow}
and described below.

\SubSection{Modeling and Simulation}
\label{sec:Modeling}

Model engineers begin by defining the FL tasks that they would like to run
on a given FL population in Python. Our library enables model engineers to
declare Federated Learning and evaluation tasks using engineer-provided
TensorFlow functions. The role of these functions is to map input tensors
to output metrics like loss or accuracy. During development, model engineers
may use sample test data or other proxy data as inputs. When deployed, the
inputs will be provided from the on-device example store via the FL runtime.

The role of the modeling infrastructure is to enable model engineers to focus
on their model, using our libraries to build and test the corresponding FL
tasks. FL tasks are validated against engineer-provided test data and
expectations, similar in nature to unit tests. FL task tests are ultimately
required in order to deploy a model as described below in
Sec.~\ref{sec:Versioning}.

The configuration of tasks is also written in Python and includes runtime
parameters such as the optimal number of devices in a round as well as model
hyperparameters like learning rate. FL tasks may be defined in groups: for
example, to evaluate a grid search over learning rates. When more than
one FL task is deployed in an FL population, the FL service chooses among them
using a dynamic strategy that allows alternating between training and evaluation
of a single model or A/B comparisons between models.

Initial hyperparameter exploration is sometimes done in simulation using proxy
data. Proxy data is similar in shape to the on-device data but drawn from a
different distribution -- for example, text from Wikipedia may be viewed
as proxy data for text typed on a mobile keyboard. Our modeling tools
allow deployment of FL tasks to a simulated FL server and a fleet of cloud
jobs emulating devices on a large proxy dataset. The simulation executes the
same code as we run on device and communicates with the server using simulated
FL populations. Simulation can scale to a large number of devices and is
sometimes used to pre-train models on proxy data before it is refined by FL
in the field.

\SubSection{Plan Generation}
\label{sec:PlanGen}

Each FL task is associated with an FL plan. Plans are automatically
generated from the combination of model and configuration supplied by
the model engineer.
Typically, in data center training, the information which is encoded
in the FL plan would be represented by a Python program which
orchestrates a TensorFlow graph. However, we do not execute Python
directly on the server or devices. The FL plan's purpose is to describe the desired
orchestration independent of Python.

An FL plan consists of two parts: one for the device and one for the
server. The device portion of the FL plan contains, among other things:
the TensorFlow graph itself, selection criteria for
training data in the example store, instructions on how to batch data and
how many epochs to run on the device, labels for the nodes in the graph which
represent certain computations like loading and saving weights, and so
on. The server part contains the aggregation logic, which is encoded
in a similar way. Our libraries automatically split the
part of a provided model's computation which runs on device from the
part that runs on the server (the aggregation).

\SubSection{Versioning, Testing, and Deployment}
\label{sec:Versioning}

Model engineers working in the federated system are able to work productively
and safely, launching or ending multiple experiments per day. But because each
FL task may potentially be RAM-hogging or incompatible with version(s) of
TensorFlow running on the fleet, engineers rely on the FL system's versioning,
testing, and deployment infrastructure for automated safety checks.

An FL task that has been translated into an FL plan is not accepted by the
server for deployment unless certain conditions are met. First, it must
have been built from auditable, peer reviewed code. Second, it must have bundled test
predicates for each FL task that pass in simulation. Third, the resources
consumed during testing must be within a safe range of expected resources
for the target population. And finally, the FL task tests must pass on every
version of the TensorFlow runtime that the FL task claims to support,
as verified by testing the FL task's plan in an Android emulator.

\textit{Versioning} is a specific challenge for on-device machine learning. In
contrast to data-center training, where the TensorFlow runtime and graphs can
generally be rebuilt as needed, devices may be running a version of the
TensorFlow runtime that is many months older than what is required by the FL plan
generated by modelers today. For example, the old runtime may be missing a
particular TensorFlow operator, or the signature of an operator may have changed
in an incompatible way. The FL infrastructure deals with this problem by
generating \textit{versioned} FL plans for each task. Each versioned FL plan
is derived from the default (unversioned) FL plan by transforming its
computation graph to achieve compatibility with a deployed TensorFlow version.
Versioned and unversioned plans must pass the same release tests, and are
therefore treated as semantically equivalent.
We encounter about one incompatible change that can be fixed with a graph
transformation every three months, and a slightly smaller number that cannot be
fixed without complex workarounds.

\SubSection{Metrics}

As soon as an FL task has been accepted for deployment, devices checking in
may be served the appropriate (versioned) plan. As soon as an FL round closes,
that round's aggregated model parameters and metrics are written to the
server storage location chosen by the model engineer.

Materialized model metrics are annotated with additional data, including
metadata like the source FL task's name, FL round number within
the FL task, and other basic operational data. The metrics themselves are
summaries of device reports within the round via approximate order statistics
and moments like mean. The FL system provides analysis tools for model engineers
to load these metrics into standard Python numerical data science packages
for visualization and exploration.

\Section{Applications}
\label{sec:applications}

Federated Learning applies best in situations where the on-device data
is more relevant than the data that exists on servers (e.g., the
devices generate the data in the first place), is privacy-sensitive,
or otherwise undesirable or infeasible to transmit to servers. Current
applications of Federated Learning are for supervised learning tasks, typically
using labels inferred from user activity (e.g., clicks or typed words).

\begin{description}[style=unboxed,leftmargin=0cm]
\item[On-device item ranking]
A common use of machine learning in mobile applications is
selecting and ranking items from an on-device inventory. For example,
apps may expose a search mechanism for information retrieval
or in-app navigation, for example settings search on Google Pixel devices
\citep{googleai18settingssearch}. By ranking these results on-device, expensive
calls to the server (in e.g., latency, bandwidth or power consumption
dimensions) are eliminated, and any potentially private information from
the search query and user selection remains on the device. Each user
interaction with the ranking feature can become a labeled
data point, since it's possible to observe the user's interaction with the
preferred item in the context of the full ranked list.

\item[Content suggestions for on-device keyboards]
On-device keyboard implementations can add value to users by suggesting
relevant content -- for example, search queries that are related to the input
text. Federated Learning can be used to train ML models
for triggering the suggestion feature, as well as ranking the items
that can be suggested in the current context. This approach has been taken
by Google's Gboard mobile keyboard team, using our FL system \citep{yang18gboardquery}.

\item[Next word prediction] Gboard also used our FL platform to train
  a recurrent neural network (RNN) for next-word-prediction
  \citep{gboard}.  This model, which has about 1.4 million parameters,
  converges in 3000 FL rounds after processing 6e8 sentences from
  1.5e6 users over 5 days of training (so each round takes about 2--3
  minutes).\footnote{This is roughly $7\times$ slower than in
    comparable data center training of the same model. However, we do
    not believe this type of comparison is the primary one -- our main
    goal is to enable training on data that is not available in the
    data center.  In fact, for the model mentioned different proxy
    data was used for data center training. Nevertheless, fast
    wall-clock convergence time is important for enabling model
    engineers to iterate rapidly, and hence we are continuing to
    optimize both our system and algorithms to decrease convergence
    times.} It improves top-1 recall over a baseline $n$-gram model
  from 13.0\% to 16.4\%, and matches the performance of a
  server-trained RNN which required 1.2e8 SGD steps. In live A/B
  experiments, the FL model outperforms both the $n$-gram and the
  server-trained RNN models.

\end{description}

\Section{Operational Profile}
\label{sec:Evaluation}

In this section we provide a brief overview of some key operational
metrics of the deployed FL system, running production workloads for
over a year; Appendix \ref{app:op} provides additional details. These
numbers are examples only, since we have not yet applied FL to a
diverse enough set of applications to provide a complete
characterization. Further, all data was collected in the process of
operating a production system, rather than under controlled conditions
explicitly for the purpose of measurement. Many of the performance
metrics here depend on the device and network speed (which can vary by
region); FL plan, global model and update sizes (varies per
application); number of samples per round and computational complexity
per sample.

We designed the FL system to elastically scale with the number and
sizes of the FL populations, potentially up into the
billions. Currently the system is handling a cumulative FL population
size of approximately 10M daily active devices, spanning several
different applications.

\begin{figure}[htb]
  \centering
  \includegraphics[width=\columnwidth]{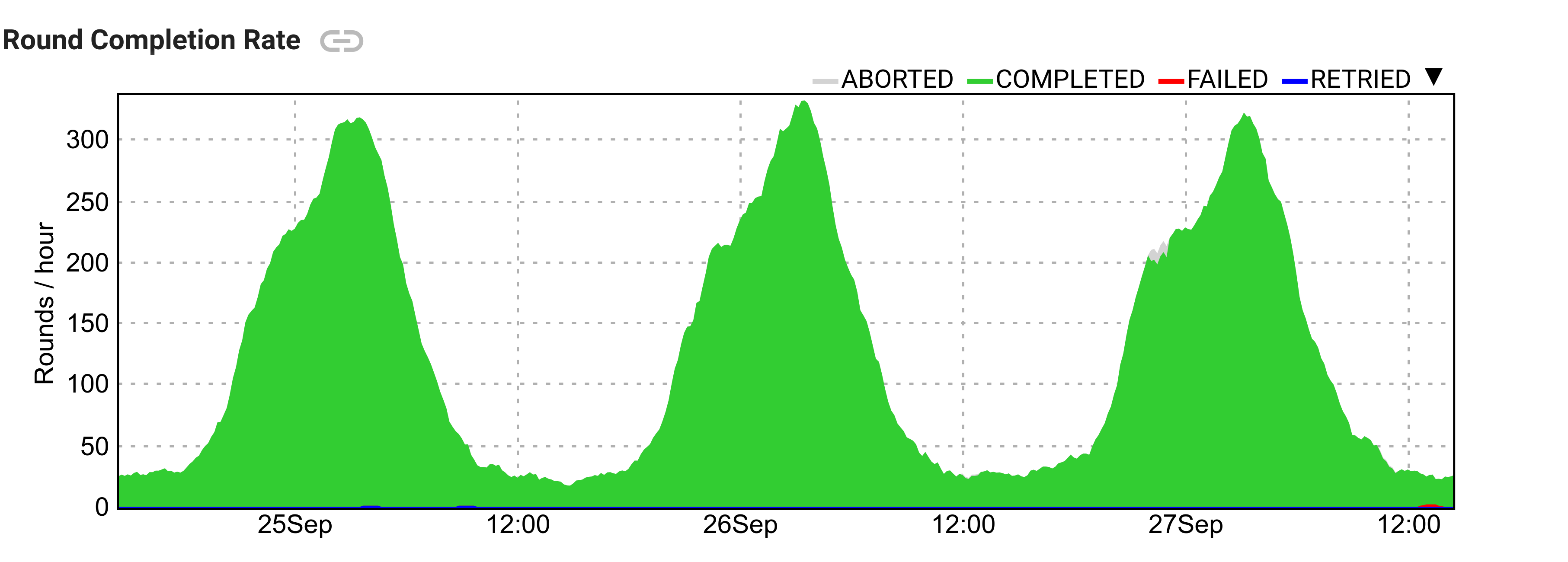}
  \caption{Round Completion Rate}
  \label{fig:CompletionRate}
\end{figure}

As discussed before, at any point in time only a subset of devices
connect to the server due to device eligibility and pace steering.
Given this, in practice we observe that up to 10k devices are
participating simultaneously. It is worth noting that the number of
participating devices depends on the (local) time of day (see
Fig.~\ref{fig:CompletionRate}). Devices are more likely idle and
charging at night, and hence more likely to participate. We have observed a
$4\times$ difference between low and high numbers of participating
devices over a 24 hours period for a US-centric population.

Based on the previous work of \citet{FL4} and experiments we have conducted
on production FL populations, for most models receiving updates from a
few hundred devices per FL round is sufficient (that is, we see
diminishing improvements in the convergence rate from training on
larger numbers of devices). We also observe that on average the
portion of devices that drop out due to computation errors, network
failures, or changes in eligibility varies between 6\% and
10\%. Therefore, in order to compensate for device drop out as well as
to allow stragglers to be discarded, the server typically selects
130\% of the target number of devices to initially participate. This
parameter can be tuned based on the empirical distribution of device
reporting times and the target number of stragglers to ignore.

\Section{Related Work}
\label{sec:Related}

\begin{description}[style=unboxed,leftmargin=0pt]

\item[Alternative Approaches]

To the best of our knowledge, the system we described is the first
production-level Federated Learning implementation, focusing primarily
on the Federated Averaging algorithm running on mobile phones. Nevertheless,
there are other ways to learn from data stored on mobile phones, and other
settings in which FL as a concept could be relevant.

In particular, \citet{pihur2018differentially} proposes an algorithm
that learns from users' data without performing aggregation on the
server and with additional formal privacy guarantees.  However, their
work focuses on generalized linear models, and argues that their
approach is highly scalable due to avoidance of synchronization and
not requiring to store updates from devices.  Our server design
described in Sec.~\ref{server}, rebuts the concerns about scalability
of the synchronous approach we are using, and in particular shows that
updates can be processed online as they are received without a need to
store them.  Alternative proposals for FL algorithms include
\citet{smith2017federated, kamp2018efficient}, which would be on the
high-level compatible with the system design described here.

In addition, Federated Learning has already been proposed in the context of
vehicle-to-vehicle communication \citep{samarakoon2018federated} and medical
applications \citep{brisimi2018federated}. While the system described in this
work as a whole does not directly apply to these scenarios, many aspects of
it would likely be relevant for production application.

\citet{takayuki18client_selection} focuses on applying FL in different
environmental conditions, namely where the server can reach any subset of
heterogeneous devices to initiate a round, but receives updates sequentially due
to cellular bandwidth limit. The work offers a resource-aware selection algorithm
maximizing the number of participants in a round, which is implementable within our
system.

\item[Distributed ML]

There has been significant work on distributed machine learning, and large-scale
cloud-based systems have been described and are used in practice. Many
systems support multiple distribution schemes, including \textit{model
  parallelism} and \textit{data parallelism}, e.g., \citet{dean12large}
and \citet{GRAPH_LAB}. Our system imposes a more structured approach
fitting to the domain of mobile devices, which have much lower
bandwidth and reliability compared to datacenter nodes. We do not
allow for arbitrary distributed computation but rather focus on a
synchronous FL protocol. This domain specialization allows us, from
the system viewpoint, to optimize for the specific use case.

A particularly common approach in the datacenter is the
\textit{parameter server}, e.g., \citet{PARAM_SERVER, dean12large,
  TF}, which allows a large number of workers to collaborate on a
shared global model, the parameter vector. Focus in that line of work
is put on an efficient server architecture for dealing with vectors of
the size of $10^9$ to $10^{12}$. The parameter server provides global
state which workers access and update asynchronously.  Our approach
inherently cannot work with such a global state, because we require a
specific rendezvous between a set of devices and the FL server to
perform a synchronous update with Secure Aggregation.

\item[MapReduce]

For datacenter applications, it is now commonly accepted that
MapReduce \citep{MAP_REDUCE} is not the right framework for ML
training. For the problem space of FL, MapReduce is a close
relative. One can interpret the FL server as the Reducer, and FL
devices as Mappers.  However, there are also fundamental technical
differences compared to a generic MapReduce framework. In our system,
FL devices own the data on which they are working. They are fully
self-controlled actors which attend and leave computation rounds at
will. In turn, the FL server actively scans for available FL devices,
and brings only selected subsets of them together for a round of
computation. The server needs to work with the fact that many devices
drop out during computation, and that availability of FL devices
varies drastically over time. These very specific requirements are
better dealt with by a domain specific framework than a generic
MapReduce.

\end{description}

\vspace{-10pt}
\Section{Future Work}
\label{sec:Future}

\begin{description}[style=unboxed,leftmargin=0pt]

\item[Bias]

The Federated Averaging \citep{FL4} protocol assumes that all devices
are equally likely to participate and complete each round. In
practice, our system potentially introduces bias by the fact that
devices only train when they are on an unmetered network and
charging. In some countries the majority of people rarely have access
to an unmetered network. Also, we limit the deployment of our device
code only to certain phones, currently with recent Android versions
and at least 2 GB of memory, another source of potential bias.

We address this possibility in the current system as follows: During
FL training, the models are not used to make user-visible predictions;
instead, once a model is trained, it is evaluated in live A/B
experiments using multiple application-specific metrics (just as with
a datacenter model). If bias in device participation or other issues
lead to an inferior model, it will be detected at this point. So far,
we have not observed this to be an issue in practice, but this is
likely application and population dependent.
Further quantification of these possible effects across a wider set of
applications, and if needed algorithmic or systems approaches to
mitigate them, are important directions for future work.

\item[Convergence Time]

We noted in Sec.~\ref{sec:applications} that we currently
observe a slower convergence time for Federated Learning compared to ML
on centralized data where training is backed by the power of a data
center. Current FL algorithms such as Federated Averaging can only efficiently
utilize 100s of devices in parallel, but many more are available; FL would
greatly benefit from new algorithms that can utilize increased parallelism.

On the operational side, there is also more which can be done. For
example, the time windows to select devices for training and wait for
their reporting is currently configured statically per FL population. It
should be dynamically adjusted to reduce the drop out rate and increase
round frequency. We should ideally use online ML for tuning this and
other parameters of the protocol configuration, bringing in e.g. time
of the day as context.

\item[Device Scheduling]

Currently, our multi-tenant on-device scheduler uses a simple worker
queue for determining which training session to run next (we avoid
running training sessions on-device in parallel because of their high
resource consumption). This approach is blind to aspects like which
apps the user has been frequently using. It's possible for us to end up
repeatedly training on older data (up to the expiration date) for some apps,
while also neglecting training on newer data for the apps the user is frequently
using. Any optimization here, though, has to be carefully evaluated
against the biases it may introduce.

\item[Bandwidth]

When working with certain types of models, for example recurrent
networks for language modeling, even small amounts of raw data can
result in large amounts of information (weight updates) being
communicated. In particular, this might be more than if we would just
upload the raw data. While this could be viewed as a tradeoff for
better privacy, there is also much which can be improved. To reduce
the bandwidth necessary, we implement compression techniques such as those of
\citet{BANDWIDTH} and \citet{konecny18expanding}.
In addition to that, we can modify the training
algorithms to obtain models in quantized representation
\citep{QUANTIZATION}, which will have synergetic effect with bandwidth
savings and be important for efficient deployment for inference.

\item[Federated Computation]

We believe there are more applications besides ML for the general
device/server architecture we have described in this paper. This is also
apparent from the fact that this paper contains no explicit
mentioning of any ML logic. Instead, we refer abstractly to 'plans', 'models',
'updates' and so on.

We aim to generalize our system from Federated Learning to
\textit{Federated Computation}, which follows the same basic
principles as described in this paper, but does not restrict
computation to ML with TensorFlow, but general MapReduce like
workloads. One application area we are seeing is in \textit{Federated
  Analytics}, which would allow us to monitor aggregate device
statistics without logging raw device data to the cloud.

\end{description}



\vspace{-14pt}
\section*{Acknowledgement}

Like most production scale systems, there are many more contributors
than the authors of this paper. The following people, at least, have
directly contributed to design and implementation: \ackn.

\clearpage

\clearpage
\appendix

\Section{Operational Profile Data}
\label{app:op}

In this section we present operational profile data for one of the FL
populations that are currently active in the deployed FL system,
augmenting the discussion in Sec.~\ref{sec:Evaluation}. The subject FL
population primarily comes from the same time zone.

\begin{figure}[htp]
  \centering
  \includegraphics[width=\columnwidth]{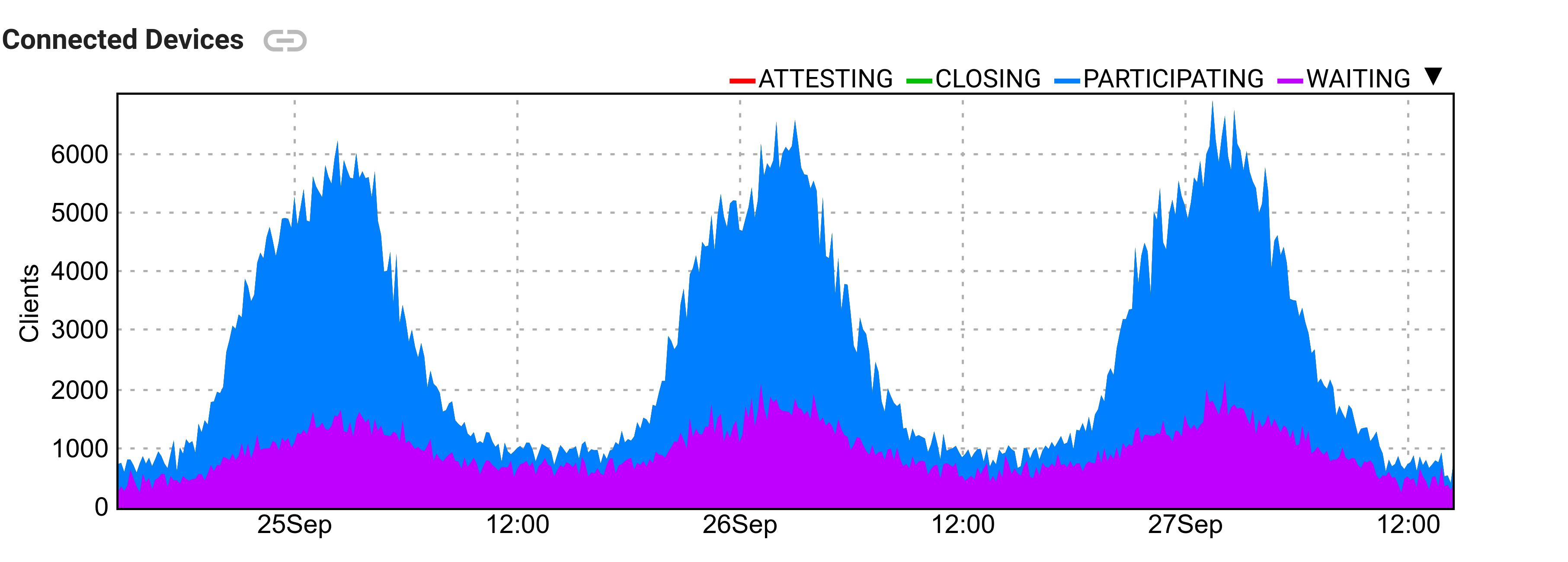}
  \includegraphics[width=\columnwidth]{round_completion_rate.png}
  \caption{A subset of the connected devices over three days (top) in states
    ``participating'' (blue) and ``waiting'' (purple). Other states
    (``closing'' and ``attesting'') are too rare to be visible in this graph.
    The rate of successful round completions (green, bottom) is also shown,
    along with the rate of other outcomes (``failure'', ``retry'', and
    ``abort'') plotted on the same graph but too low to be visible.}
  \label{fig:ConnectedDevicesCompletionRate}
\end{figure}

Fig.~\ref{fig:ConnectedDevicesCompletionRate} illustrates how
availability of the devices varies through the day and its impact on
the round completion rate. Because the FL server schedules an FL task
for execution only once a desired number of devices are available and
selected, the round completion rate oscillates in sync with device
availability.

\begin{figure}[htb]
  \includegraphics[width=\columnwidth]{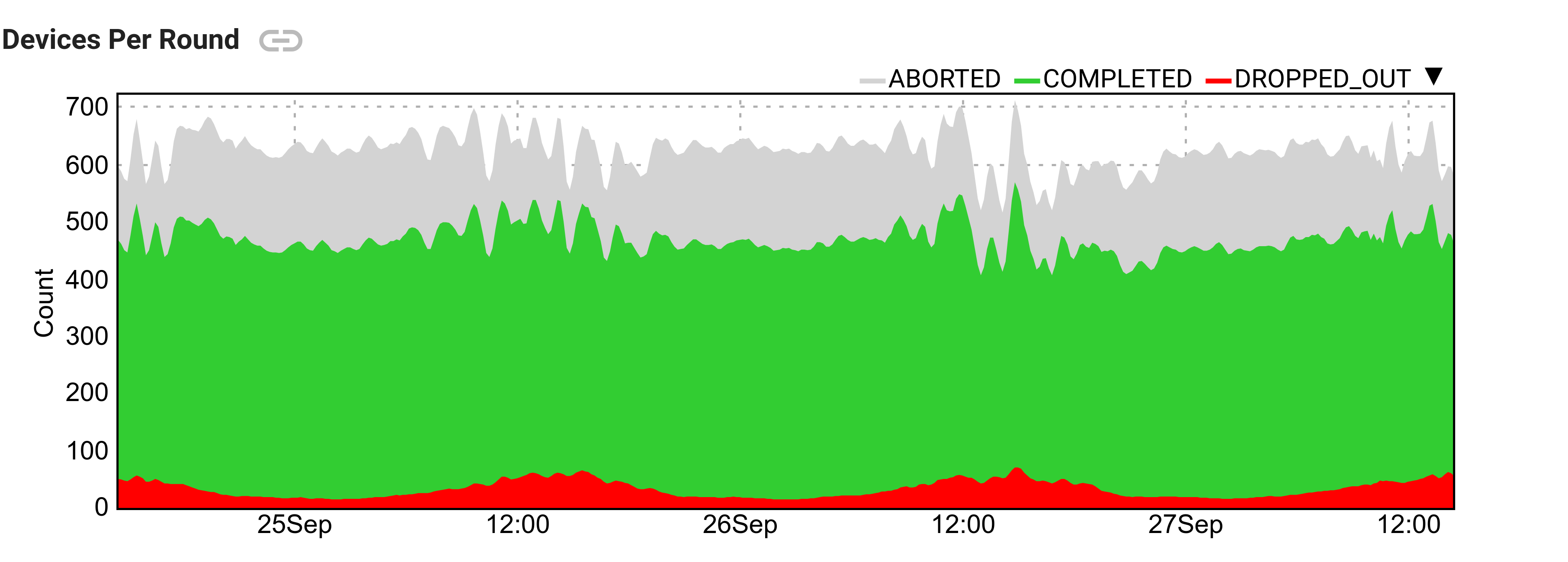}
  \caption{Average number of devices completed, aborted and dropped out from
  round execution}
  \label{fig:DevicesPerRound}
\end{figure}

Fig.~\ref{fig:DevicesPerRound} illustrates the average number of devices
participating in an FL task round and the outcomes of the participation. Note
that in each round the FL server selects more devices for the participation than
desired to complete to offset the devices that drop out during execution.
Therefore in each round there are devices that were aborted after a desired
number of devices successfully complete. Another noteworthy aspect is drop out
rate correlation with the time of day, specifically the drop out rate is higher
during the day time compared to the night time. This is explained by higher
probability of the device eligibility criteria changes due interaction with
a device.

\begin{figure}[htb]
  \includegraphics[width=\columnwidth]{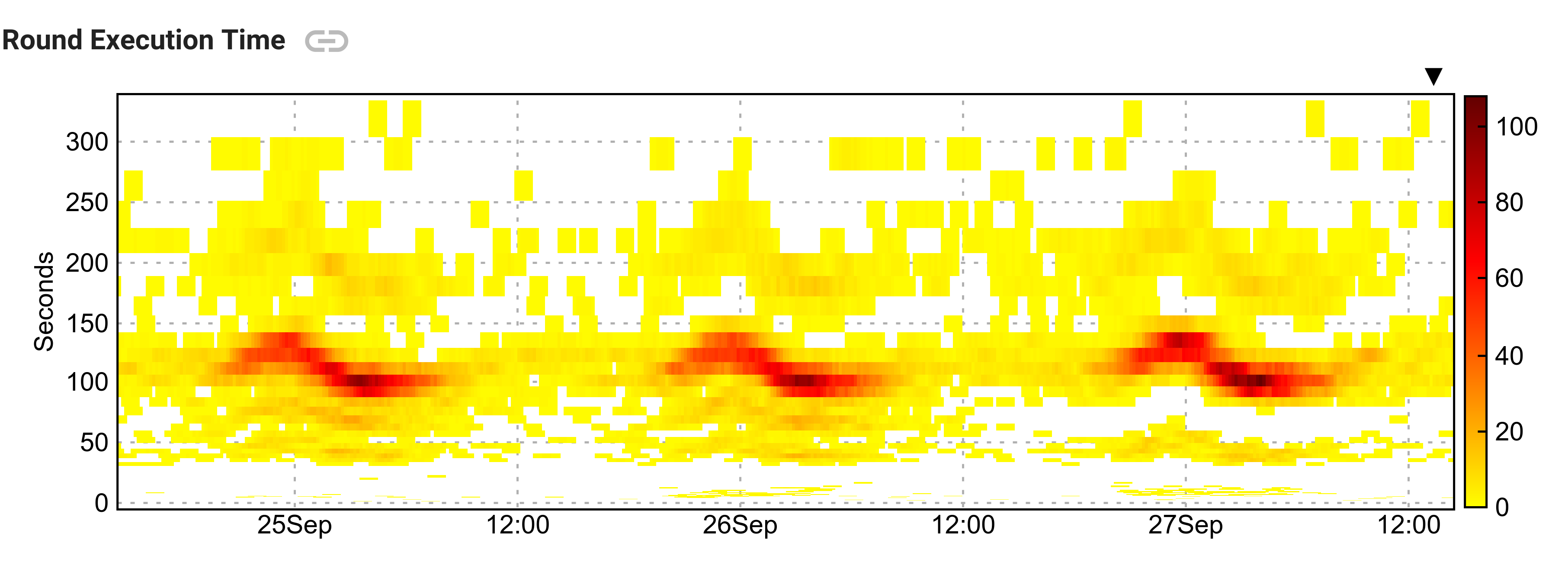}
  \includegraphics[width=\columnwidth]{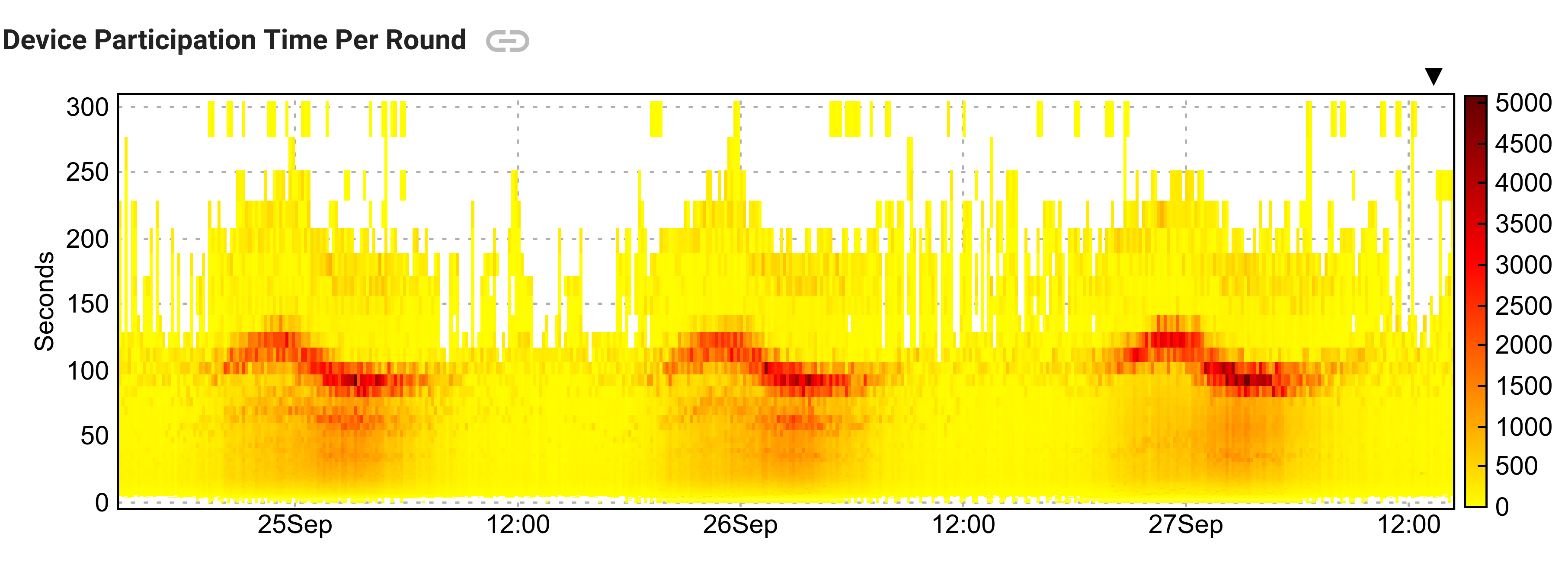}
  \caption{Round execution and device participation time}
  \label{fig:RoundExecutionTimeDeviceParticipationTime}
\end{figure}

Fig.~\ref{fig:RoundExecutionTimeDeviceParticipationTime} shows distribution
of round run and device participation time. There are two noteworthy
observations. First is that the round run time is roughly equal to the majority
of the device participation time which is explained by the fact that the
FL server selects more than needed devices for participation and stops execution
when enough devices complete. Second is that device participation time is
capped. This is a mechanism used by the FL server to deal with straggler
devices; i.e., the round run time capped by the server.

\begin{figure}[htb]
  \includegraphics[width=\columnwidth]{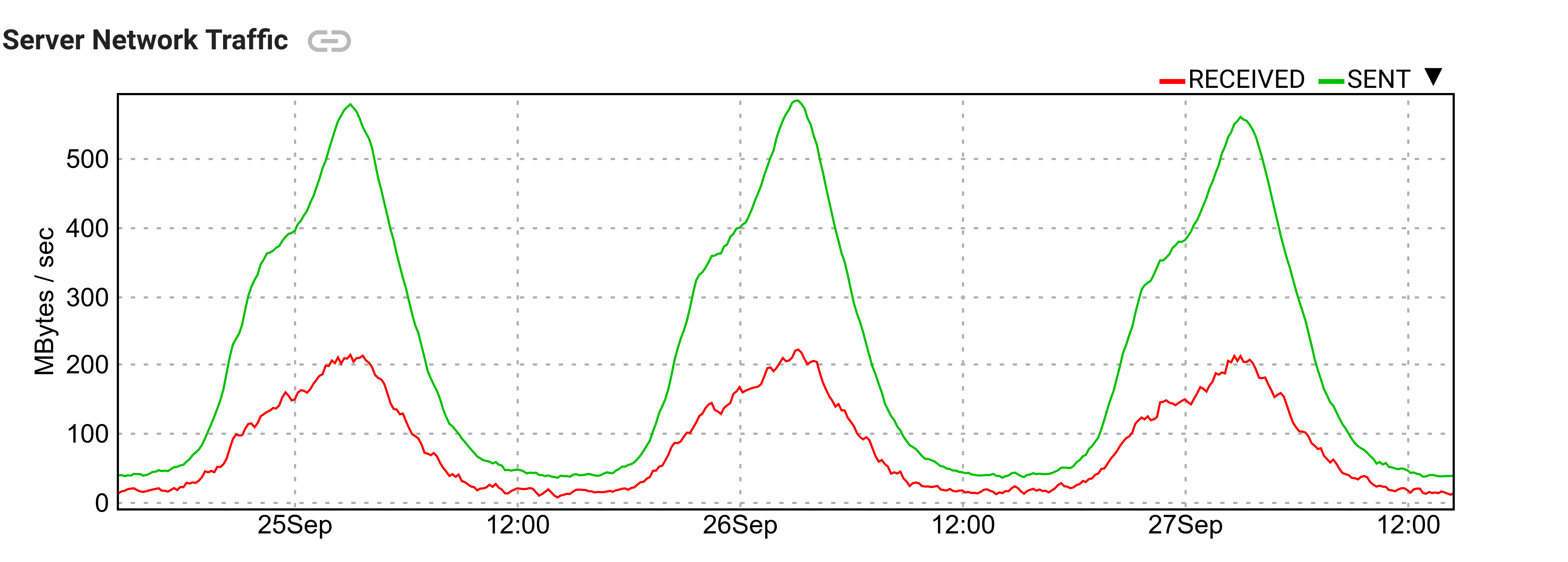}
  \caption{Server network traffic}
  \label{fig:ServerNetworkTraffic}
\end{figure}

Fig.~\ref{fig:ServerNetworkTraffic} illustrates the asymmetry in server network
traffic, specifically that download from server dominates upload. There are
several aspects that contribute. Namely each device downloads both an FL task
plan and current global model (plan size is comparable with the global model)
whereas it uploads only updates to the global model; the model updates
are inherently more compressible compared to the global model.

Tab.~\ref{tab:TrainingRoundVisualizations} shows the training round
session shape visualizations generated from the clients' training
state event logs. As shown, 75\% of clients complete their training
rounds successfully, 22\% of clients complete their training rounds
but have their results rejected by the server (these are the devices
which report back after the reporting window already closed), and 2\%
of clients are interrupted before being able to complete their round
(e.g., because the device exited the idle state).

\begin{table}[]
  \centering
  \begin{tabular}{lrr}
    \hline
    Session Shape               & Count     & Percent \\ \hline
    -v{[}{]}+\textasciicircum{} & 1,116,401 & 75\%    \\
    -v{[}{]}+\#                 & 327,478   & 22\%    \\
    -v{[}!                      & 29,771    & 2\%     \\ \hline
  \end{tabular}
  \caption{Distribution of on-device training round sessions. Legend: -~=~FL
  server checkin, v~=~downloaded plan, [~=~training started, ]~=~training
  completed, +~=~upload started, \textasciicircum{}~=~upload completed,
  \#~=~upload rejected, !~=~interrupted.}
  \label{tab:TrainingRoundVisualizations}
\end{table}

\clearpage
\Section{Federated Averaging}
\label{app:fedavg}

In this section, we show the Federated Averaging algorithm from \citet{FL4}
for the interested reader.

\newcommand{\winit}{w_{\text{init}}}
\newcommand{\T}{\rule{0pt}{2.2ex}}  
\newcommand{\comm}[1]{ \ \ \  // \emph{#1}}
\newcommand{\bcomm}[1]{// \emph{#1}}
\newcommand{\nclients}{K}

\begin{algorithm}[h]
\begin{algorithmic}
\SUB{Server executes:}
   \STATE initialize $w_0$
   \FOR{each round $t = 1, 2, \dots$}
     \STATE Select $1.3\nclients$ eligible clients to compute updates
     \STATE Wait for updates from $\nclients$ clients (indexed $1, \dots, \nclients)$
     \STATE $(\T\Delta^k, n^k) = \text{ClientUpdate}(w)$ from client $k \in [\nclients]$.
     \STATE $\T\bar{w}_t = \sum_k \Delta^k$ \comm{Sum of weighted updates}
     \STATE $\T\bar{n}_t = \sum_k n^k$ \comm{Sum of weights}
     \STATE $\T\Delta_t = \Delta^k_t / \bar{n}_t$ \comm{Average update}
     \STATE $w_{t+1} \leftarrow w_t + \Delta_t$
   \ENDFOR
   \STATE

 \SUB{ClientUpdate($w$):}\
 \STATE $\mathcal{B} \leftarrow$ (local data divided into minibatches)
 \STATE $n \leftarrow |\mathcal{B}|$  \comm{Update weight}
 \STATE $\winit \leftarrow w$
 \FOR{batch $b \in \mathcal{B}$}
    \STATE $w \leftarrow w - \eta \grad \loss(w; b)$
  \ENDFOR
 \STATE $\Delta \leftarrow n \cdot (w - \winit)$ \comm{Weighted update}
 \STATE \bcomm{Note $\Delta$ is more amenable to compression than $w$}
 \STATE return $(\Delta,  n)$ to server
\end{algorithmic}
\caption{\fedavglong targeting updates from $\nclients$ clients per
  round.}\label{alg:fedavg}
\end{algorithm}

\end{document}